%% file: PaperForReview.tex
\documentclass[10pt,twocolumn,letterpaper]{article}

\usepackage{cvpr}              % To produce the CAMERA-READY version
\usepackage{graphicx}
\usepackage{amsmath}
\usepackage{amssymb}
\usepackage{booktabs}
\usepackage{bbm}
\usepackage{multirow}
\usepackage{wrapfig}
\usepackage{bm}

\usepackage[pagebackref,breaklinks,colorlinks]{hyperref}

\usepackage[capitalize]{cleveref}
\crefname{section}{Sec.}{Secs.}
\Crefname{section}{Section}{Sections}
\Crefname{table}{Table}{Tables}
\crefname{table}{Tab.}{Tabs.}

\def\cvprPaperID{2349} % *** Enter the CVPR Paper ID here
\def\confName{CVPR}
\def\confYear{2022}

\begin{document}

\title{Many-to-many Splatting for Efficient Video Frame Interpolation}

\author{Ping Hu$^{1}$\thanks{Work primarily done while Ping was interning at Adobe.}~~~~~~~~~~~Simon Niklaus$^2$~~~~~~~~~~~Stan Sclaroff$^1$~~~~~~~~~~~Kate Saenko$^{1,3}$\\
 $^1$Boston University~~~~~$^2$Adobe Research~~~~~$^3$MIT-IBM Watson AI Lab\\
}
\maketitle

%%%%%%%%% ABSTRACT
\begin{abstract}
Motion-based video frame interpolation commonly relies on optical flow to warp pixels from the inputs to the desired interpolation instant.
Yet due to the inherent challenges of motion estimation (e.g. occlusions and discontinuities), most state-of-the-art interpolation approaches require subsequent refinement of the warped result to generate satisfying outputs, which drastically decreases the efficiency for multi-frame interpolation.
In this work, we propose a fully differentiable Many-to-Many (M2M) splatting framework to interpolate frames efficiently.
Specifically, given a frame pair, we estimate multiple bidirectional flows to directly forward warp the pixels to the desired time step, and then fuse any overlapping pixels. 
In doing so, each source pixel renders multiple target pixels and each target pixel can be synthesized from a larger area of visual context.
This establishes a many-to-many splatting scheme with robustness to artifacts like holes.
Moreover, for each input frame pair, M2M only performs motion estimation once and has a minuscule computational overhead when interpolating an arbitrary number of in-between frames, hence achieving fast multi-frame interpolation.
We conducted extensive experiments to analyze M2M, and found that it significantly improves the efficiency while maintaining high effectiveness.
\end{abstract}

%%%%%%%%% BODY TEXT

\input{introduction}
\input{figures/overview}
\input{related}

\input{method}

\input{experiments}

\section{Conclusion}
In this work, we present a many-to-many splatting technique to efficiently interpolate intermediate video frames. 
We first design a motion refinement network to generate multiple sub-motion vectors for each pixel. 
These sub-motion fields are then applied to forward warp the pixels to any desired time step, which are then fused to obtain the final output.
By sharing the computation for the flow refinement and only requiring little compute to generate each frame, our method is especially well-suited for multi-frame interpolation.
Experiments on multiple benchmark datasets demonstrate that the proposed method achieves effectiveness with superior efficiency.

%\noindent{\textbf{Acknowledgements.Acknowledgements.}} This work was supported in part by DARPA and NSF.

%%%%%%%%% REFERENCES
{\small
\bibliographystyle{ieee_fullname}
\bibliography{egbib}
}

\end{document}

%% file: introduction.tex
\section{Introduction}
\label{sec:intro}
Video frame interpolation (VFI) aims to increase frame rates of videos by synthesizing intermediate frames in between the original ones~\cite{siyao2021deep,baker2011database}.
As a classic problem in video processing, VFI contributes to many practical applications, including slow-motion animation~\cite{jiang2018super}, video editing~\cite{meyer2018deep}, video compression~\cite{wu2018video}, \etc. 
In recent years, a plethora of techniques for video frame interpolation have been proposed~\cite{meyer2018phasenet,meyer2015phase,zhang2020video,liu2020enhanced,Yu_2021_ICCV,tulyakov2021time,reda2019unsupervised}.  However, frame interpolation remains an unsolved problem due to challenges like occlusions, large motion, and lighting changes.

\input{figures/tesear}

The referenced research can roughly be categorized into motion-free and motion-based, depending on whether or not cues like optical flow are incorporated~\cite{kroeger2016fast,sun2018pwc}.
Motion-free models typically rely on kernel prediction~\cite{cheng2020video,ding2021cdfi,niklaus2021revisiting,peleg2019net} or spatio-temporal decoding~\cite{choi2020channel,choi2021motion,kalluri2020flavr}, which are effective but limited to interpolating frames at fixed time steps and their runtime increases linearly in the number of desired output frames.
%incur major computational costs for each individual output.
On the other end of the spectrum, motion-based approaches establish dense correspondences between frames and apply warping to render the intermediate pixels.  

A common motion-based technique estimates bilateral flow for the desired time step and then synthesizes the intermediate frame via backward warping~\cite{huang2020rife,bao2019depth,park2021asymmetric,park2020bmbc,jiang2018super}. 
The estimation of bilateral motion is challenging though and incorrect flows can easily degrade the interpolation quality.
As a result,  for each time step, these methods typically apply a synthesis network to refine the bilateral flows.
Another motion-based solution is to forward warp pixels to the desired time step 
via optical flow~\cite{baker2011database}.
However, forward warping is subject to holes and ambiguities where multiple pixels map to the same location. 
Therefore, image refinement networks are commonly adopted to correct remaining artifacts~\cite{niklaus2020softmax,niklaus2018context,xue2019video}. 
However, both of these approaches require significant amounts of compute, and the refinement networks need to be executed for each of the desired interpolation instants.
This decreases their efficiency in multi-frame interpolation tasks since their runtime increases linearly in the number of desired output frames.

\input{figures/warp}
We address these challenges and strive for efficiency with a Many-to-Many (M2M) splatting framework. 
Specifically, our proposed M2M splatting estimates multiple bidirectional flow fields and then efficiently forward warps the input images to the desired time step before fusing any overlapping pixels.
Since we directly operate on  pixel colors, the quality and resolution of the underlying optical flow play a critical role. 
For this reason, we first apply an off-the-shelf optical flow estimator~\cite{kroeger2016fast,sun2018pwc} to
extract the inter-frame motion between the two input frames at a coarse level.
Based on this low-resolution optical flow estimate, a Motion Refinement Network (MRN) predicts multiple flow vectors for each pixel at the full-resolution which we then use for our image synthesis through many-splatting.

Conventional motion-based frame interpolation methods only estimate one inter-frame motion vector for each pixel~\cite{niklaus2020softmax,niklaus2018context,xue2019video,huang2020rife,bao2019depth,park2021asymmetric,park2020bmbc}.
However and as shown in Fig.~\ref{fig:warp} (a), forward warping with such a motion field manifests as many-to-one splatting, leaving unnecessary holes in the warped result.
To overcome this limitation, we model a many-to-many relationship among pixels by predicting multiple motion vectors for each of the input pixels, 
and then forward warping the pixels to multiple locations at the desired time step. 
As shown in Fig.~\ref{fig:warp} (b), many-to-many splatting allows for more complex interactions among pixels, \ie each source pixel is allowed to render multiple target  pixels and each target pixel can be synthesized with a larger area of visual context.
Unsurprisingly, many-to-many splatting leads to many more overlapping pixels.  
To merge these, we further introduce a learning-based fusion strategy which adaptively combines pixels that map to the same location.

Since the optical flow estimation step  in our pipeline predicts time-invariant correspondence estimates, it only needs to be performed once for a given input frame pair. 
Once the many-to-many inter-frame motion has been established, generating new in-between frames only requires warping and fusing the input images. This is in stark contrast to previous approaches that leverage refinement networks~\cite{niklaus2020softmax,niklaus2018context}, allowing us to perform multi-frame interpolation an order of magnitude faster as shown in Fig.~\ref{fig:tesear}.
%The warping and fusion operations are associated with each of the desired time steps, yet has a very slight overhead since we directly operate in the full-resolution pixel color domain, avoiding the need of post-processing steps like extra deep image synthesis/refinement networks~\cite{niklaus2020softmax,niklaus2018context}. 
%As a result, M2M effectively and efficiently interpolates multiple and arbitrary intermediate frames for videos. 
%We conduct extensive experiments on multiple datasets, and as shown in Fig.~\ref{fig:tesear}, M2M  improves the interpolation efficiency while maintaining state-of-the-art accuracy. 

In summary, we propose 1) a Motion-Refinement Network that estimates a many-to-many relationship between the two input images, 2) a learning-based pixel fusion strategy which resolves ambiguities between overlapping pixels, and 3) a well-motivated Many-to-Many (M2M) splatting synthesis model for efficient and effective frame interpolation. Our experiments demonstrate that M2M  achieves high effectiveness with fast speed, e.g. $\sim$40 ms/f using a Titan X to perform $\times$8 interpolation of 2K videos.

%% file: figures/tesear.tex
\begin{figure}[t]
    \centering
    \includegraphics[width=1\linewidth]{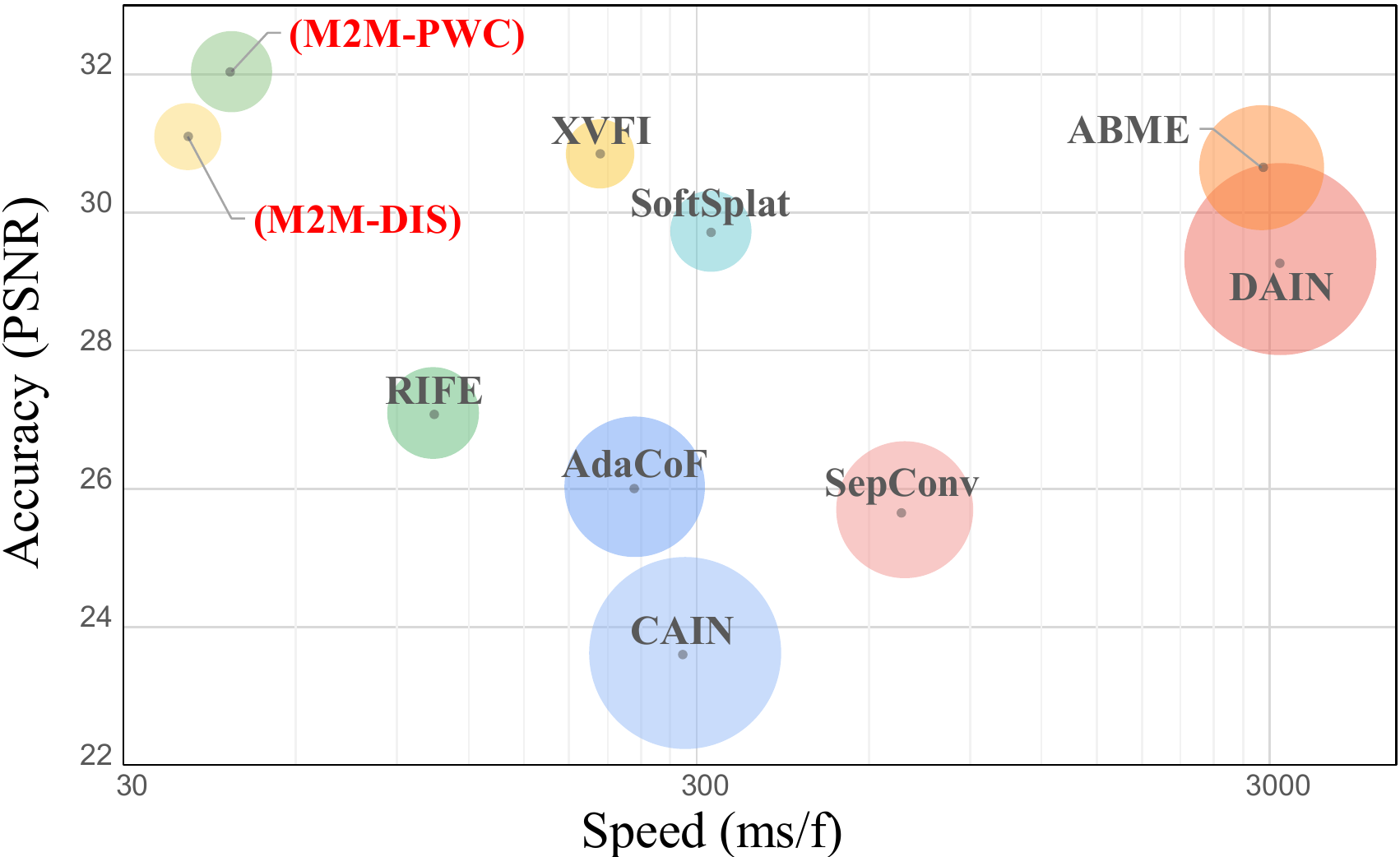}
    \vspace{-0.6cm}
    \caption{\small{Performance for $\times8$ interpolation on a ``2K'' version of X-TEST~\cite{sim2021xvfi}.  Runtimes for all methods were measured using a Titan X GPU.  The size of each circle indicates the number of model parameters. Results for related methods include RIFE~\cite{huang2020rife}, SoftSplat~\cite{niklaus2020softmax}, AdaCof~\cite{lee2020adacof}, SepConv~\cite{niklaus2017sepconv}, XVFI~\cite{sim2021xvfi}, DAIN~\cite{bao2019depth}, ABME~\cite{park2021asymmetric}, and CAIN~\cite{park2020bmbc}.
    We evaluate our proposed M2M splatting using two different off-the-shelf flow estimators, ``PWC'' denoting PWC-Net~\cite{sun2018pwc} and ``DIS'' denoting DISFLow~\cite{kroeger2016fast}.}}
    \label{fig:tesear}
    \vspace{-0.1cm}
\end{figure}

%% file: figures/warp.tex
\begin{figure}[t]
    \centering
    \includegraphics[width=0.95\linewidth]{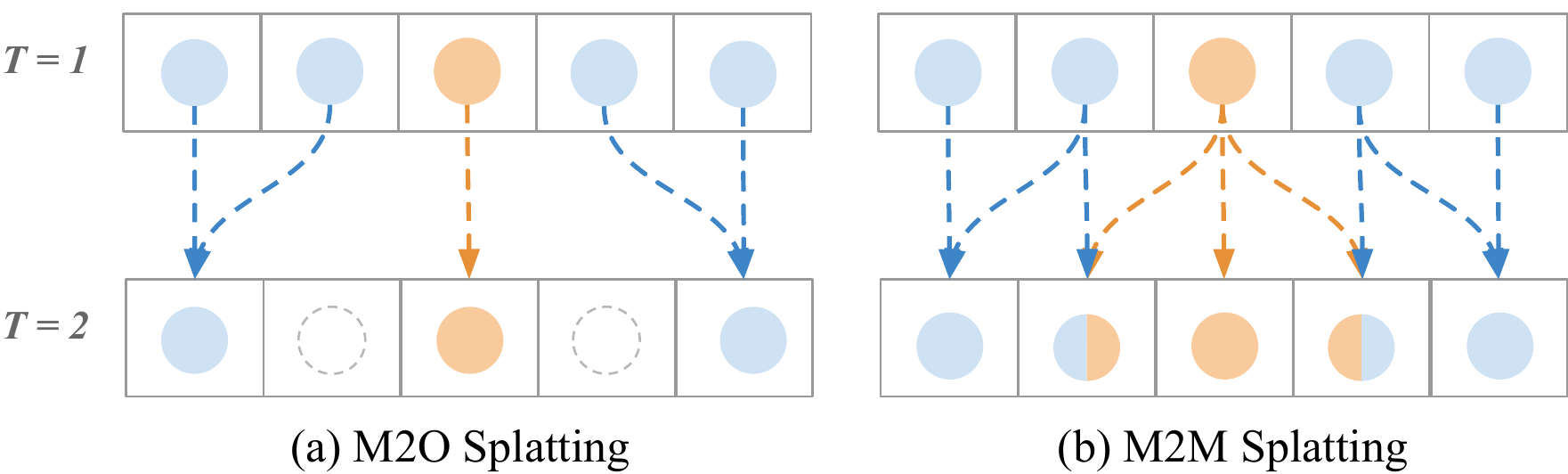}
    \vspace{-0.15cm}
    \caption{\small{(a) Many-to-one splatting versus (b) many-to-many splatting for zooming motion in a scene containing blue and orange pixels. M2O splatting may results in holes, while M2M splatting allows for a more flexible image formation model.}}
    \label{fig:warp}
    \vspace{-0.1cm}
\end{figure}

%% file: figures/overview.tex
\begin{figure*}[t]
    \centering
    \includegraphics[width=\linewidth]{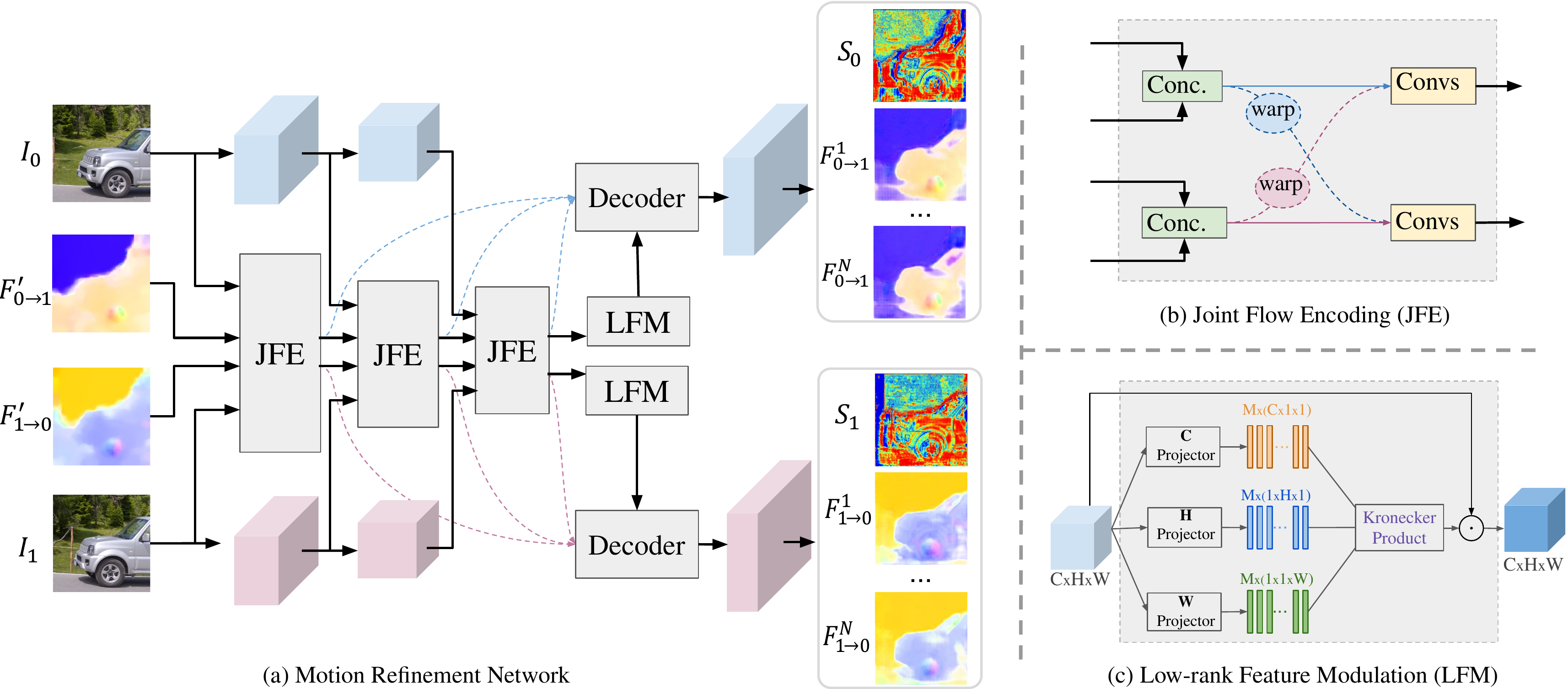}
    \vspace{-0.2cm}
    \caption{\small{ Overview of the  (a) Motion Refinement Network and its core modules: (b) Joint Flow Encoding  and (c) Low-rank Feature Modulation. Given an image pair $\{I_0,I_1\}$ and the initial bidrectional inter-frame flow $\{F'_{0\rightarrow1}, F'_{1\rightarrow0}\}$, the goal is to generate multiple refined bidirectional flows $\{F^i_{0\rightarrow1}, F^i_{1\rightarrow0}\}_{i=1}^N$ and the color reliability maps $\{S_{0},S_{1}\}$. The ``warp'' in the JFE denotes backward warping.}}
    \label{fig:overview}
    \vspace{-0.1cm}
\end{figure*}

%% file: related.tex
\section{Related Work}
\label{sec:related}
%Video frame interpolation (VFI) is one of the fundamental tasks for computer vision and video processing. 
%Along with the developments in deep learning techniques, the area of VFI has witnessed a substantial advancement in the past a few years. 
%In this section, we will discuss recent techniques of this topic. 

%Motion plays a key role for existing VFI methods, as it explicitly models pixel-level correspondence and trajectories across frames~\cite{baker2011database}.
Motion-based video frame interpolation approaches typically estimate optical flows~\cite{kroeger2016fast,sun2018pwc} from given frames, and then propagate pixels/features  to the desired target time step~\cite{xue2019video,yuan2019zoom,zhang2020flexible,niklaus2022splatting}. 
Forward warping is an efficient solution to achieve this goal~\cite{baker2011database}. 
With bidirectional optical flow between given frames, Niklaus~\etal~\cite{niklaus2018context} directly forward warp the images as well as contextual features to the interpolation instant before utilizing a synthesis network to render the output frame. To make this splatting fully differentiable, they further introduce softmax splatting~\cite{niklaus2020softmax} which allows them to train the feature extraction end-to-end.
Splatting has its downsides though, since it is not only necessary to address ambiguities of multiple pixels mapping to the same location but it is also necessary to handle the holes that are present in the sparse result.

To avoid having to handle these challenges, some methods are based on backward warping instead~\cite{bao2019memc,sim2021xvfi}. 
The necessary bilateral flow can, for example, be approximated from off-the-shelf flow estimates through a neural network~\cite{jiang2018super} or depth-based splatting~\cite{bao2019depth}. 
Park~\etal~\cite{park2020bmbc,park2021asymmetric} extend these ideas and introduce a network to further improve the motion representations while Huang~\etal~\cite{huang2020rife} learn to directly estimate bilateral flows.
However, estimating bilateral flow is still challenging and the backward warped pixels may still suffer from artifacts. 
As a result, these methods also rely on image synthesis networks to improve the interpolation quality~\cite{huang2020rife,park2020bmbc,park2021asymmetric,niklaus2020softmax,niklaus2018context}. Though shown to be effective, the bilateral flow estimation and the image synthesis networks need to be fully executed for each desired output, leading to a linearly increasing runtime when interpolating more than one in-between frame.

In contrast to these methods, our M2M approach relies on many-to-many splatting to address the issues with forward warping without relying on an image synthesis network or bilateral flow approximation/estimation.

Another dominant research direction for VFI aims to avoid explicit motion estimation altogether. One popular approach is to resample input pixels with spatially adaptive filters~\cite{liu2017video,peleg2019net}. Niklaus~\etal~\cite{niklaus2017adaconv} estimate spatially-varying kernels which in subsequent work are decomposed into separable kernels~\cite{niklaus2017sepconv,niklaus2021revisiting}, which also formulate a many-to-many correlations between pixels. However, as local patches suffer from a limited spatial range, deformable convolutions are introduced to handle large motion~\cite{lee2020adacof,cheng2020video}.
To improve model efficiency, Ding~\etal~\cite{ding2021cdfi} introduce model compression~\cite{lee2020adacof}.
Spatio-temporal decoding methods are also proposed to directly convert spatio-temporal features into target frames via channel attention~\cite{choi2020channel,choi2021motion} or 3D convolutions~\cite{kalluri2020flavr}.
However, most of these methods  generate outputs at a fixed time, typically halfway between the input images, which limits arbitrary-time interpolation and linearly increases the runtime for multi-frame interpolation.     
%In contrast, our method effectively interpolate frames of arbitrary numbers at arbitrary time steps with very slight extra computation.

%% file: method.tex
\section{Many-to-many Splatting Framework}
\label{sec:method}
In this section, we describe our Many-to-Many (M2M) splatting framework for video frame interpolation. 
Given an input frame pair,  we first estimate the bidirectional motion with an off-the-shelf method~\cite{sun2018pwc,kroeger2016fast}. 
A Motion Refinement Network (Fig.~\ref{fig:overview} (a)) then takes the off-the-shelf motion predictions as input and estimates multiple motion vectors as well as a reliability score for each individual pixel in the input frames.
Lastly, all input pixels are forward warped to the desired target time step several times via each of the multiple motion vectors, and finally merged to generate the output via a pixel fusion that leverages the estimated reliability score.
With full end-to-end supervision, our M2M framework is able to achieve not only efficiency but also effectiveness.
In the following, we first present the Motion Refinement Network in Sec.~\ref{subsec:multi-flow}, then introduce the multi-splatting and fusion of pixels in Sec.~\ref{subsec:m2msplatting}.

\subsection{Motion Refinement Network}
\label{subsec:multi-flow}
Optical flow is a common technique to model inter-frame motion in videos.
Yet directly applying an off-the-shelf optical flow estimator and forward warping pixels based on this estimate may be challenging. 
Optical flow only models a single motion vector for each pixel, thus limiting the area that a pixel can splat to and thus potentially causing holes. 
Moreover, most optical flow estimators are supervised with training data at a relatively low resolution and forcing them to process high-resolution frames may yield poor results.
In contrast, we present the Motion Refinement Network (MRN) to upsample and refine an off-the-shelf optical flow estimate while predicting multiple motion vectors per pixel.
As shown in Fig.~\ref{fig:overview} (a), the MRN pipeline is composed of three parts: Motion Feature Encoding, Low-rank Feature Modulation, and Output Decoding.

\noindent\textbf{Motion Feature Encoding} aims to encode multi-stage motion features from the input frames $\{I_0, I_1\}$ as well as the optical flow $\{F'_{0\rightarrow1}, F'_{1\rightarrow0}\}$ estimated by an off-the-shelf estimator~\cite{sun2018pwc,kroeger2016fast} at a coarse resolution.
As outlined in Fig.~\ref{fig:overview} (a), the encoding process is designed in a hierarchical manner.
At first, we extract two $L$-level image feature pyramids from $I_0$ and $I_1$, with the zeroth-level being the images themselves. 
To generate the feature representations at each pyramid level, we utilize two convolutional layers with intermittent PReLU activations to downsample the features from the previous level by a factor of two.
In our implementation, we use $L=4$, and the numbers of feature channels from shallow to deep are $16$, $32$, $64$, and $128$ respectively.

Then, from the zeroth to the last level, we apply Joint Flow Encoding (JFE) modules as illustrated in Fig.~\ref{fig:overview} (b) to progressively generate motion feature pyramids for the bidirectional flow fields $F'_{0\rightarrow1}$ and  $F'_{1\rightarrow0}$. 
In the $l$-th level's JFE module, the motion and image features from the previous level are warped towards each other. Specifically, the features from the pyramid corresponding to $I_0$ are warped towards $I_1$ and vice versa using the off-the-shelf optical flow estimates.
Then, the original features and the warped features are combined and downsampled using a two-layer CNN to encode the $l$-th level's motion features. 
% Ideally, the bidirectional flows are symmetric to each other, and our siamese network design helps to reinforce this. % as a regularization to the network thus benefiting the model's effectiveness.

\noindent\textbf{Low-rank Feature Modulation} is designed to further enhance the motion feature representations  with a low-rank constraint. 
The idea behind this module is that flow fields of natural dynamic scenes are highly structured due to the underlying physical constraints, which can be exploited by low-rank models to enhance the motion estimation quality~\cite{dong2014nonlocal,tang2020lsm,Lara_2016_CVPR,roberts2009learning}.
To avoid formulating explicit optimization objectives like in previous methods, which may be inefficient in high-resolution applications, we draw inspirations from Canonical Polyadic (CP) decomposition~\cite{kolda2009tensor} and construct an efficient low-rank modulation module to enhance each flow's feature maps with low-rank characteristics.

As shown in Fig.~\ref{fig:overview} (c), given an input feature map of size $C\times H\times W$, three groups of projectors are adopted to respectively shrink the feature maps into the \textit{channel}, \textit{height}, and \textit{width} dimensions. 
Each projector is composed of a pooling layer, $1\times 1$ conv layers, and a sigmoid function. 
We apply $M$ projectors for each of the three dimensions which results in three groups of 1-D features, whose sizes can be represented as $M\times (C\times 1\times 1)$ for the \textit{channel} dimension,  $M\times (1\times H\times 1)$  for the \textit{height} dimension, and $M\times (1\times 1\times W)$ for the \textit{width} dimension.
Then, for each of the $M$ vectors from the three dimensions, we apply the \textit{Kronecker Product} to get a rank-1 tensor, whose shape is $C\times H\times W$.
The $M$ rank-1 tensors are later averaged point-wise. 
To ensure low-rank characteristic, $M$ is set to be smaller than $C$, $H$, and $W$ (we adopt $M=16$ in this work). 
%The process is illustrated in Fig.~\ref{fig:kronkprod}.
%Since the projector contains a sigmoid function, elements of the final low-rank tensor are  $\in [0, 1]$.
We combine the input features and the low-rank tensor via point-wise multiplication, where the latter serves as weights to modulate the former with low-rank characteristics.

\input{figures/multi_flow}

\input{figures/visualwarp}
Deep learning-based low-rank constraints have also been utilized for model compression~\cite{phan2020stable}, segmentation~\cite{chen2020tensor} and  image reconstruction~\cite{zhang2021learning}. In this work we explore the application to motion modeling and demonstrate its effectiveness on the task of video frame interpolation.

\noindent\textbf{Output Decoding} generates $N$ motion vectors as well as the reliability scores for each input pixel based on the motion feature pyramids and the feature maps subject to the low-rank prior. 
We adopt deconv layers to enlarge the spatial size of the feature maps.
That is, the decoder operates in $L$ stages from coarse to fine while leveraging the features encoded by the JFE modules.
At the last decoding stage, the full-resolution feature maps for the flow in each direction are converted into multiple fields $\{F^i_{0\rightarrow 1}, F^i_{1\rightarrow 0}\}_{i=1}^N$ as well as the corresponding reliability maps $\{S_{0}, S_{1}\}$, which are later utilized to fuse pixels that map to the same location when generating the new in-between frames. An example of these outputs is visualized in Fig.~\ref{fig:multi_flow}.

\subsection{Pixel Warping and Fusion}
\label{subsec:m2msplatting}
%In this subsection, we present the Many-to-many (M2M) Splatting to interpolate arbitrary intermediate frames.
The previously estimated multi-motion fields are first used to forward warp pixels to a given target time step. 
Later, we present a fusion strategy to combine the colors of overlapping pixels in the output. 
Since both the warping and fusion steps operate with pixels' colors without any subsequent post-processing steps, an intermediate frame can be interpolated with minuscule computational overhead.

\noindent\textbf{Pixel Warping.} So far, we have generated $N$ full-resolution bidirectional  motion fields $\{F^n_{0\rightarrow1},F^n_{1\rightarrow0}\}_{n=1}^N$ and pixel-wise reliability scores $\{S_0,S_1\}$ for the input video frame pair $\{I_0,I_1\}$.
The next step is to synthesize an intermediate frame $I_t$ at the desired time step $t\in(0,1)$.
Under the assumption of linear motion, we first scale each pixel's motion vectors by the desired interpolation time $t$ as:
\begin{equation}
\label{eq:flow2t}
\begin{aligned}
        F^n_{0\rightarrow t}(i_0) &= t \cdot F^n_{0\rightarrow 1}(i_0) \\
        F^n_{1\rightarrow t}(i_1) &= (1-t) \cdot F^n_{1\rightarrow 0}(i_1)
\end{aligned}
\end{equation}

\noindent where $i_0$ and $i_1$ denote the $i$-th source pixel in $I_0$ and $I_1$ respectively. Then, a source pixel $i_s$ is forward warped by its $n$-th motion vector to $i_{s\rightarrow t}^n = \phi_F(i_s,F^n_{s\rightarrow t})$ at the desired intermediate time $t$, with $s\in\{0,1\}$ representing the source frame, $\phi_F$ is the forward warping operation, and $F^n_{s\rightarrow t}$ is the $n$-th sub-motion vector of $i_s$ as defined in Eq.~\ref{eq:flow2t}.
    
We first consider utilizing a single motion vector for warping, which means each pixel is only warped to one location in the target frame. In dynamic scenes, the motion vectors may overlap with each other thus resulting in a many-to-one (M2O) propagation where the pixel set after fusion is smaller than the actual pixel set of frame. This results in holes as shown in Fig.~\ref{fig:visual_warp} (a). 
Though exploiting multiple source frames lessens this issue, M2O warping still restricts each source pixel to only render a small 4-pixel vicinity in the output frame.
This limits the effectiveness in representing and thus interpolating regions with complex interactions among the pixels, as shown in Fig.~\ref{fig:visual_warp} (b).

Fortunately, such limitations can be alleviated through many-to-many (M2M) pixel splatting by using multiple motion vectors to model the motion of each source pixel. 
We forward warp each pixel in the source $s$ with $N$ ($N>1$) sub-motion vectors to $t$, and get the set of warped pixels,
\begin{align}
\hat{I}_{s\rightarrow t} = \bigcup\limits_{n=1}^N\hat{I}^n_{s\rightarrow t}
\end{align}
Many-to-many splatting relaxes the restriction that each source pixel can only contribute to a single location. 
Therefore it allows the underlying motion estimator to learn to reason about occlusions, and model complex color interactions across a larger area of pixels. 
 
\noindent\textbf{Pixel Fusion.}
By applying M2M warping to all the input pixels in $\{I_0, I_1\}$, we get the complete warped pixel set where multiple target pixels may correspond to the same pixel locations:  $\hat{I}_t =\hat{I}_{0\rightarrow t}\bigcup\hat{I}_{1\rightarrow t}$.
To fuse warped pixels overlap with each other, we measure each of the pixels' importance from three aspects: the temporal relevance, brightness consistency, and the reliability score.

\textit{1) Temporal Relevance} \textbf{r}{$_i$} characterizes changes not based on motion (e.g. lighting changes)  between a source frame and the target. 
For simplicity, we adopt linear interpolation by setting $r_i=1-t$ if $i$ comes from $I_0$ and $r_i=t$ otherwise, with $t$ being the desired interpolation time. 

\textit{2) Brightness Consistency} \textbf{b}{$_i$} indicates occlusions by comparing a frame to its target through backward warping:
\begin{align}
b_i=
\begin{cases}
-1\cdot||I_0(i)-I_1(i+F_{0\rightarrow 1}(i))||_1, &\text{\hspace{-0.2cm}if}~~i\in I_0,\\
-1\cdot||I_1(i)-I_0(i+F_{1\rightarrow 0}(i))||_1, &\text{\hspace{-0.2cm}if}~~i\in I_1,
\end{cases}
\label{eqn:occ}
\end{align}

The effectiveness of Eq.~\ref{eqn:occ} is not decided only by the motion but also by the pixels' colors, which can be affected by various factors like noise, ambiguous appearance, and changes in shading~\cite{niklaus2020softmax,baker2011database}. To enhance the robustness, we thus further adopt a learned per-pixel reliability score.

\textit{3) Reliability Score} \textbf{s}$_i$ is jointly estimated together with the motion vectors through the  Motion Refinement Network as introduced in Sec.~\ref{subsec:multi-flow} and learned from data. 

With these three measurements, we fuse the overlapped pixels at a location $j$ in the form of weighted summation,
\begin{align}
    I_t(j) = \frac{\sum_{i\in\hat{I}_t}\mathbbm{1}_{i=j}\cdot e^{(b_{i}\cdot s_{i}\cdot\alpha)}\cdot r_{i}\cdot c_i}{\sum_{i\in\hat{I}_t}\mathbbm{1}_{i=j}\cdot  e^{(b_{i}\cdot s_{i}\cdot\alpha)}\cdot r_{i}}
\end{align}
\noindent where $c_i$ represents the $i$-th warped pixel's original color, $\alpha$ is a learnable parameter adjusting the scale of weights, $\hat{I}_t$ is the set of all the warped pixels at time $t$, and $\mathbbm{1}_{i=j}$ indicates if the warped pixel  $i$ is mapping to the pixel location $j$.

We note that our final fusion function is similar to SoftSplat~\cite{niklaus2020softmax} in the form of softmax weighting, however our method differs  in three aspects. First, we provide a solution to directly operate in the pixel color domain, while SoftSplat splats features and utilizes an image synthesis network instead. Second, we propose a general framework for fusing pixels from multiple frame, while SoftSplat fuses each frame individually. Third, we introduce the learning based reliability score to fuse overlapping pixels in a data-driven manner while SoftSplat uses feature consistency.

%We design M2M to independent to specific time steps, hence it accepts any valid time steps as target and achieves arbitrary frame interpolation.
%Since  the major operations in M2M are warping and point-wise matrix multiplication, the interpolation process takes very slight computation overhead. 
%Moreover, for each frame pair the Multi-Flow refinement step need to be computed only once and can be shared for all the in-between, our method benefit with high efficiency especially in multi-frame interpolation tasks.

%% file: figures/multi_flow.tex
\begin{figure}[t]
    \centering
    \includegraphics[width=\linewidth]{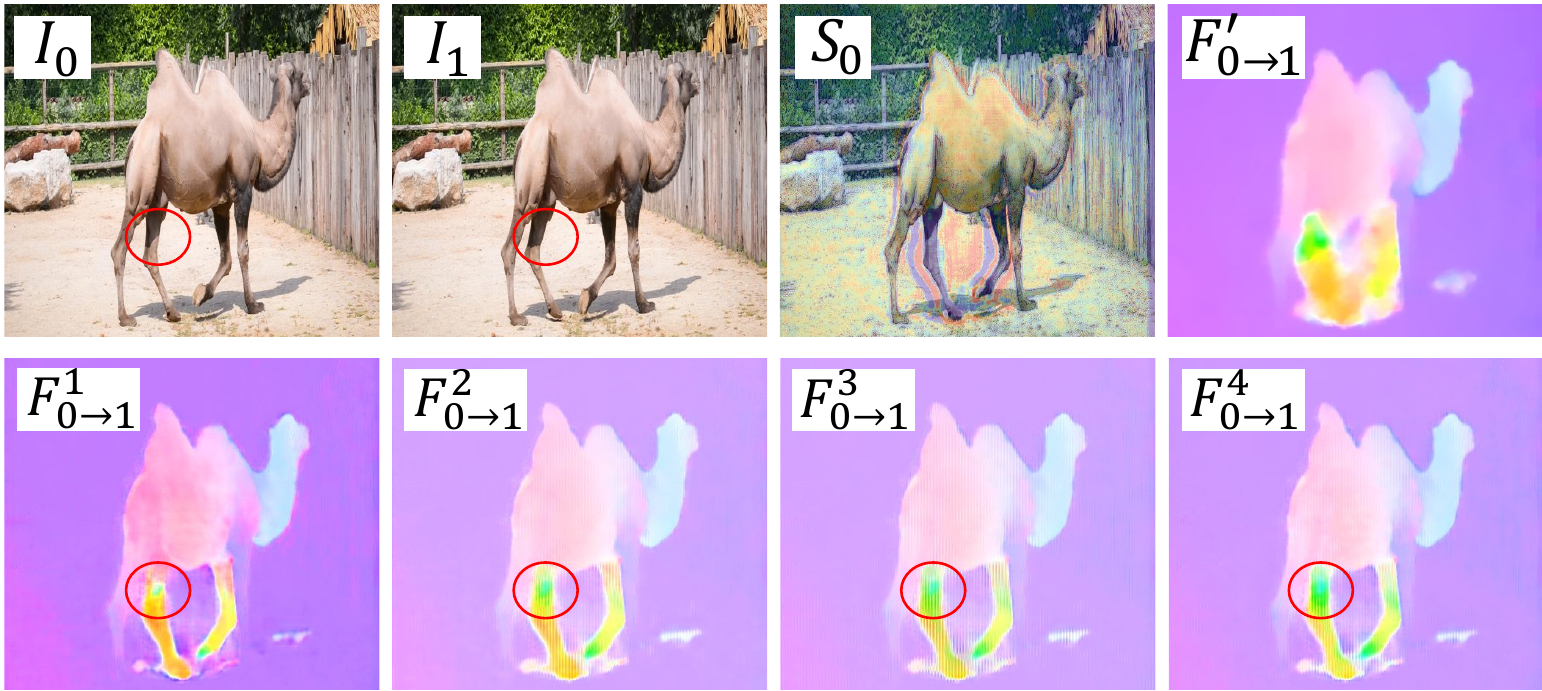}
    \vspace{-0.5cm}
    \caption{\small{Examples of the MRN's output ($N=4$). $S_0$ shows low reliability in areas with complex motion as intuitively expected. $\{F_{0\rightarrow 1}^N\}_{n=1}^4$ refine the initial flow $F'_{0\rightarrow 1}$ with better details, and decompose complex motion with shade changes (as indicated by the red circle) into multiple motion fields.
    }}
    \label{fig:multi_flow}
    \vspace{-0.1cm}
\end{figure}

%% file: figures/visualwarp.tex
\begin{figure*}[t]
    \centering
    \includegraphics[width=1\linewidth]{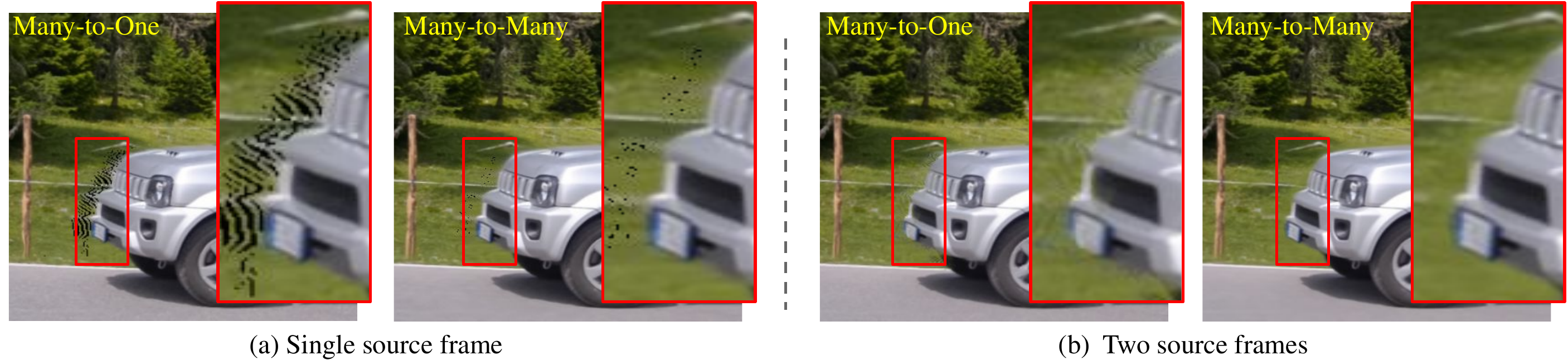}
    \vspace{-0.7cm}
    \caption{\small{Visualization of forward warping via many-to-one (M2O)  splatting and many-to-many (M2M) splatting. (a) With one source frame, M2M splatting suffers less from banding artifacts and provides improved robustness to ambiguities near the boundaries of discontinuous motion. (b) Banding artifacts can be alleviated with multiple source frames, yet M2O splatting still suffers from stray effects at boundaries due to its image formation model that is less flexible than M2M splatting. Best viewed when zoomed in. }}
    \vspace{-0.3cm}
    \label{fig:visual_warp}
\end{figure*}

%% file: experiments.tex
\input{tables/benchmark_s}
\section{Experiments}
In the section, we subsequently compare our proposed proposed to related state-of-the-art frame interpolation techniques and analyze it quantitatively as well as qualitatively.
\subsection{Datasets}
We supervise our proposed approach on the training split of Vimeo90K and test it on various datasets summarized as follows: 1)~Vimeo90K~\cite{xue2019video}, the test split containing 3,782 triplets at a resolution of 448$\times$256 pixels. 2)~UCF101~\cite{soomro2012ucf101}, a dataset containing human action videos of size 256$\times$256 pixels. A set of 379 triplets were selected by Liu~\etal~\cite{liu2017video} as a test set for frame interpolation. 3)~Xiph~\cite{xphi1994}, as proposed by Niklaus~\etal~\cite{niklaus2020softmax} where ``Xiph-2K'' is generated by downsampling 4K footage, and ``Xiph-4k'' is based on center-cropped 2K patches. 4)~ATD12K~\cite{li2020video}, containing 2,000 triplets from various animation videos at a resolution of 960$\times$480 pixels. 5)~X-TEST~\cite{sim2021xvfi}, the test set from X4K1000FPS~\cite{sim2021xvfi}, containing 15 scenes extracted from 4K videos at 1000fps. We denote the original resolution as X-TEST(4K), and additionally adopt X-TEST(2K) by downsampling X-TEST(4K) by a factor of two.

\subsection{Training}
We train our proposed pipeline in an end-to-end manner. 
Given an output $I_t$ and the ground truth $I_t^{gt}$, we define the training loss as the sum of the Charbonnier loss~\cite{charbonnier1994two} and the census loss~\cite{meister2018unflow}, $L= L_{char}+L_{cen}$.
To train the model, we utilize the 51,312 triplets from the training split of Vimeo90K~\cite{xue2019video}. 
We apply random data augmentations including spatial and temporal flipping, color jittering, and random cropping with 256$\times$256 patches.
We adopt Adam~\cite{loshchilov2018fixing} for optimization, with a weight decay of 1e-4. 
We train the model for 400k iterations with a batch size of 8, during which the learning rate is decayed from 1e-4 to 0 via cosine annealing. 
All experiments are implemented with PyTorch, and executed on a single Nvidia Titan X.

\subsection{Comparison with State-of-the-art}
We report two variants of our proposed approach based on different methods for estimating the off-the-shelf motion vectors. 
``M2M-PWC'' is based on PWC-Net~\cite{sun2018pwc}. 
In this setting, we jointly optimize PWC-Net during training and generate initial flows at 1/4 of the original resolution. 
The other variant is based on DISFlow~\cite{kroeger2016fast} and denoted as ``M2M-DIS''.
In our experiments, we generate $N$=4 sub-motion vectors for each pixel.
For comparisons, we report the performance of recent VFI approaches including: SepConv~\cite{niklaus2017sepconv}, DAIN~\cite{bao2019depth} CAIN~\cite{choi2020channel}, AdaCoF~\cite{lee2020adacof}, SoftSplat~\cite{niklaus2020softmax}, BMBC~\cite{park2020bmbc}, RIFE~\cite{huang2020rife},
and ABME~\cite{park2021asymmetric}.

We first analyze the computational costs of these models in Tab.~\ref{tab:benchmark_s}. 
We denote the required compute that is independent from the desired frame rate as ``share'', and  ``unshare''  otherwise.  
Hence the total computational complexity for interpolating $n$ frames can be calculated through ``$\#$share+$n\cdot\#$unshare''.
Motion-free methods (including SepConv, CAIN, and AdaCof) and pure bilateral-motion-based methods (like RIFE and ABME) have no share compute (denoted as ``N/A'') and their computational complexity increases linearly in the number of desired frames. 
%Moreover, these methods may have difficulty in interpolating frames of arbitrary time steps.
Approaches like SoftSplat, and BMBC can interpolate arbitrary frames, yet still suffer from both high compute and $unshare$ compute. E.g. in the $\times$8 interpolation setting,  they take 1.6 TFLOPs, and 3.1 TFLOPs respectively.
In contrast, our M2M takes only 0.1 TFLOPs in total.
Fig.~\ref{fig:fr_vs_time} (a) compares the average runtime for different methods subjct to varying interpolation factors.  
Our method is faster than all other methods in multi-frame settings. 
For $\times$16 interpolation our method takes about 5 ms to interpolate a frame, which is around 5$\times$, 20$\times$, and 100$\times$ faster than RIFE, SoftSplat, and ABME respectively. 

\input{figures/fr_vs_time}
Taking efficiency aside, our method achieves state-of-the-art performance on multiple datasets.
The metrics for $\times$2 interpolation are presented in Tab.~\ref{tab:benchmark_s}. 
On Vimeo90K and UCF101, our M2M method is on par with the recently proposed real-time method RIFE and performs slightly worse than SoftSplat and ABME.
On Xiph-2K, our M2M method achieves slightly lower PSNR than SoftSplat, yet achieves the highest SSIM among all the methods. 
Moreover, on the animation dataset ATD12K and the high-resolution dataset Xiph-``4K'', our M2M method, especially M2M-PWC, outperforms previous methods in terms of both PSNR and SSIM. 
This demonstrates our methods' effectiveness when processing high-resolution videos and the ability to generalize across domains such as animation videos.
\input{tables/benchmark_xvfi}

We report  the results for $\times$8 interpolation on the X-TEST dataset, which contains diverse sequences with both high resolution and high frame rate, in Tab.~\ref{tab:benchmark_xvfi}. 
Our M2M method outperforms all previous methods on both the original 4K full resolution (4096$\times$2160) and the downsampled 2K resolution (2048$\times$1080) with substantial advantages in efficiency.
For the models trained with Vimeo90K, ABME achieves the second-best PSNR in both 4K and 2K settings, but it takes 2,904ms to interpolate a 2K frame which is nearly 70$\times$ slower than M2M.
To evaluate the temporal consistency, we compare the accuracy at each interpolation time step in Fig.~\ref{fig:fr_vs_time} (b).
We found that previous methods tend to deteriorate when interpolating frames that are temporally centered between the inputs, while M2M achieves a flatter and smoother curve for intermediate frames.
This shows that M2M interpolates frames with not only better quality, but also higher temporal consistency.

\subsection{Method Analysis}
\input{tables/ablation}
\noindent\textbf{Ablation of Modules.} We first analyze the effectiveness of the different components of our method in Tab.~\ref{tab:ablation}. 
We start with a single motion vector for each pixel.
The first row demonstrates that directly using the off-the-shelf flow for warping leads to sub-optimal accuracy.
As shown in the second row,  applying the refinement network without joint flow encoding (JFE) and low-rank feature modulation (LFM) can already significantly improve performance by 0.97 dB and 2.38 dB for PWC-Net and DISFlow respectively.
Further applying either JFE or LFM leads to improvements of more than  0.15 dB for both off-the-shelf flow methods.
And using both JFE and LFM helps to boost the performance to 35.15 dB and 34.78 dB, respectively.
In the last two rows, we also show the impact of the reliability scores which are generated by the refinement network and utilized for the pixel fusion. 
Without this score, the performance degrades, thus highlighting the importance of this metric in comparison to only using photoconsistency.

\input{tables/branch}

\input{figures/xtest_vis}
\noindent\textbf{Effect of Number of Flows per Pixel.}
Tab.~\ref{tab:branch} compares the effect of using different numbers of the sub-motion vectors for the M2M splatting.
When $N$=1, it reduces the warping to M2O mapping, and achieves the lowest accuracy.
When increasing $N$ to 4,  M2M improves the accuracy by more than 0.1 dB, with a very slight increment in run-time ($<$1ms). 
Also, and as shown in the last row, we noticed that further increasing the number of sub-motion vectors leads to marginal improvements. 
Fig.~\ref{fig:xtest_vis} illustrates the visual results for M2O splatting and M2M splatting.

\input{tables/resolution}

\noindent\textbf{Effect of Resolution for Initial Flow Estimation}
Our method relies on an off-the-shelf optical flow estimator to generate the initial flow. 
However, most optical flow estimation models are trained using a relatively low resolutions.
Directly applying them to estimate the flow at 2K or 4K inputs  may result in sub-optimal results.
We thus study the impact of the initial flow's resolution for interpolating high-resolution frames in Tab.~\ref{tab:resolution}. Since PWC-Net is learning-based and pre-trained on small resolutions, it is less effective at processing high-resolution frames as demonstrated by the reduced interpolation quality on 4K data.
By downsampling the input by a factor of 4 or 8, the accuracy improves significantly.
In contrast, DISFlow is not supervised and hence less susceptible to similar domain gaps. % less from the overfitting problem, and tends to achieve the best performance at the original resolution of the input frames. 
% As shown, our method can be easily adapted to different types of flow estimators.

\noindent\textbf{Discussions and Limitations.} Though our method achieves very high efficiency especially for high framerate interpolation, its accuracy on low-resolution datasets like Vimeo90K is behind several state-of-the-art methods. We believe that carefully tuning and enlarging the model capacity allows M2M to compete with these state-of-the-art methods.
The proposed method renders intermediate frames based on forward warping, which may be subject to holes in the output. 
In Fig.~\ref{fig:ave_hole}, we count the average number of remaining holes (in pixels) for different configurations on Vimeo90K.
As we can see, our M2M splatting with $N$=4 is still subject to around 0.5-pixel holes in each frame on average. However, compared to the initial single sub-motion based M2O splatting, our method has significantly decreased the number of holes.
Another limitation of our method is that the many-to-many splatting process may result in blurriness as shown in Fig.~\ref{fig:xtest_vis} (d). 
This can be addressed by further improving the fusion strategy or applying a lightweight network to refine the output. % We will further study this in future research.

\input{figures/ave_hole}

%% file: tables/benchmark_s.tex
\begin{table*}[t]
\small
    \begin{tabular}{p{1.6cm}ccp{0.55cm}cp{0.62cm}p{0.62cm}p{0.62cm}p{0.62cm}p{0.62cm}p{0.62cm}p{0.62cm}p{0.62cm}p{0.62cm}p{0.62cm}} 
        \toprule[1pt] 
                  & \multicolumn{2}{c}{GFLOPs} & \multirow{2}{0.4in}{Speed~\\~~ms/f~}  & \multirow{2}{0.4in}{Arbitrary~\\~~Interp.~}  & \multicolumn{2}{c}{Vimeo90K} & \multicolumn{2}{c}{UCF101} &\multicolumn{2}{c}{ATD12K} &\multicolumn{2}{c}{Xiph-2k} &\multicolumn{2}{c}{Xiph-``4k''} \\
                  \cmidrule(lr){2-3}\cmidrule(lr){6-7}\cmidrule(lr){8-9}\cmidrule(lr){10-11}\cmidrule(lr){12-13}\cmidrule(lr){14-15}
                  & \small{share}     &\small{unshare}      &   &     & \small{\footnotesize{PSNR}}        & \small{\footnotesize{SSIM}}         & \small{\footnotesize{PSNR}}        & \small{\footnotesize{SSIM}}       &\small{\footnotesize{PSNR}}        & \small{\footnotesize{SSIM}}        &\small{\footnotesize{PSNR}}        & \small{\footnotesize{SSIM}}        &\small{\footnotesize{PSNR}}        & \small{\footnotesize{SSIM}}       \\
        \midrule
        \footnotesize{SepConv\cite{niklaus2017sepconv}}     &\scriptsize{N/A}  &93    &~101  &         &33.79   &.970   &34.78  &.967  &27.40   &.950  &34.77  &.929    &32.06   &.880    \\
        \footnotesize{DAIN\cite{bao2019depth}}              &712               &1308    &~977    &~$\checkmark$           &34.71   &.976   &35.00  &.968  &27.38   &.955  &35.97  &.940    &\underline{33.51}   &.898  \\
        \footnotesize{CAIN\cite{choi2020channel}}           &\scriptsize{N/A}  &29    &~~47   &          &34.65   &.973   &34.98  &.969  &25.28   &.952  &35.21  &.937    &32.56   &.901  \\
        \footnotesize{AdaCoF~\cite{lee2020adacof}}          &\scriptsize{N/A}  &117     &~~36 &           &34.47   &.973   &34.90  &.968  &27.75   &.950  &34.82  &.927    &32.19   &.882   \\
        \footnotesize{SoftSplat~\cite{niklaus2020softmax}}  &95               &218    &~122  &~$\checkmark$         &\underline{36.10}   &.980   &\textbf{35.39}  &.970  &28.22   &\underline{.957}  &\textbf{36.62}  &.944    &33.60   &.901    \\
        \footnotesize{BMBC~\cite{park2020bmbc}}             &441               &376    &1213    &~$\checkmark$          &35.01   &.976   &35.15  &.969  &27.68   &.945  &~~~--     &~~~--      &~~~--      &~~~--     \\
        \footnotesize{RIFE~\cite{huang2020rife}}            &\scriptsize{N/A}  &\underline{20}    &~~\textbf{17}   &         &35.51   &.978   &35.25  &.969  &28.59   &.953  &36.15  &.962    &33.27   &.942    \\
        \footnotesize{ABME~\cite{park2021asymmetric}}       &\scriptsize{N/A}  &549    &~497 &          &\textbf{36.18}   &\textbf{.981}   &\underline{35.38}  &.970  &28.71   &\textbf{.959}  &35.18  &.964    &32.36   &.940 \\
        \midrule
        \textbf{\footnotesize{M2M-PWC}}                                             &\underline{87}            &\bm{$<1$}  &~~32  &~$\checkmark$ &35.40   &.978   &35.17  &\textbf{.970}  &\textbf{29.03}   &\textbf{.959}  &\underline{36.45}  &\textbf{.967}    &\textbf{33.93}   &\textbf{.945}            \\
        \textbf{\footnotesize{M2M-DIS}}                                             &\textbf{61}               &\bm{$<1$}  &~~\underline{28}  &~$\checkmark$          &35.06   &.976   &35.13  &.968  &\underline{28.95}   &.956  &36.14  &\underline{.965}    &33.25   &\underline{.942}   \\
        \bottomrule
    \end{tabular}
\vspace{-0.2cm}
\caption{Quantitative results on the Vimeo90K, UCF101, ATD12K, and Xiph datasets. We compute models' GFLOPs and speed based on 640$\times$480 inputs. The ``share'' denotes the part of compute independent from the desired frame rate, which is in contrast to ``unshare''. }
\vspace{-0.3cm}
\label{tab:benchmark_s}
\end{table*}

%% file: figures/fr_vs_time.tex
\begin{figure}[t]
\centering
    \includegraphics[width=1.0\linewidth]{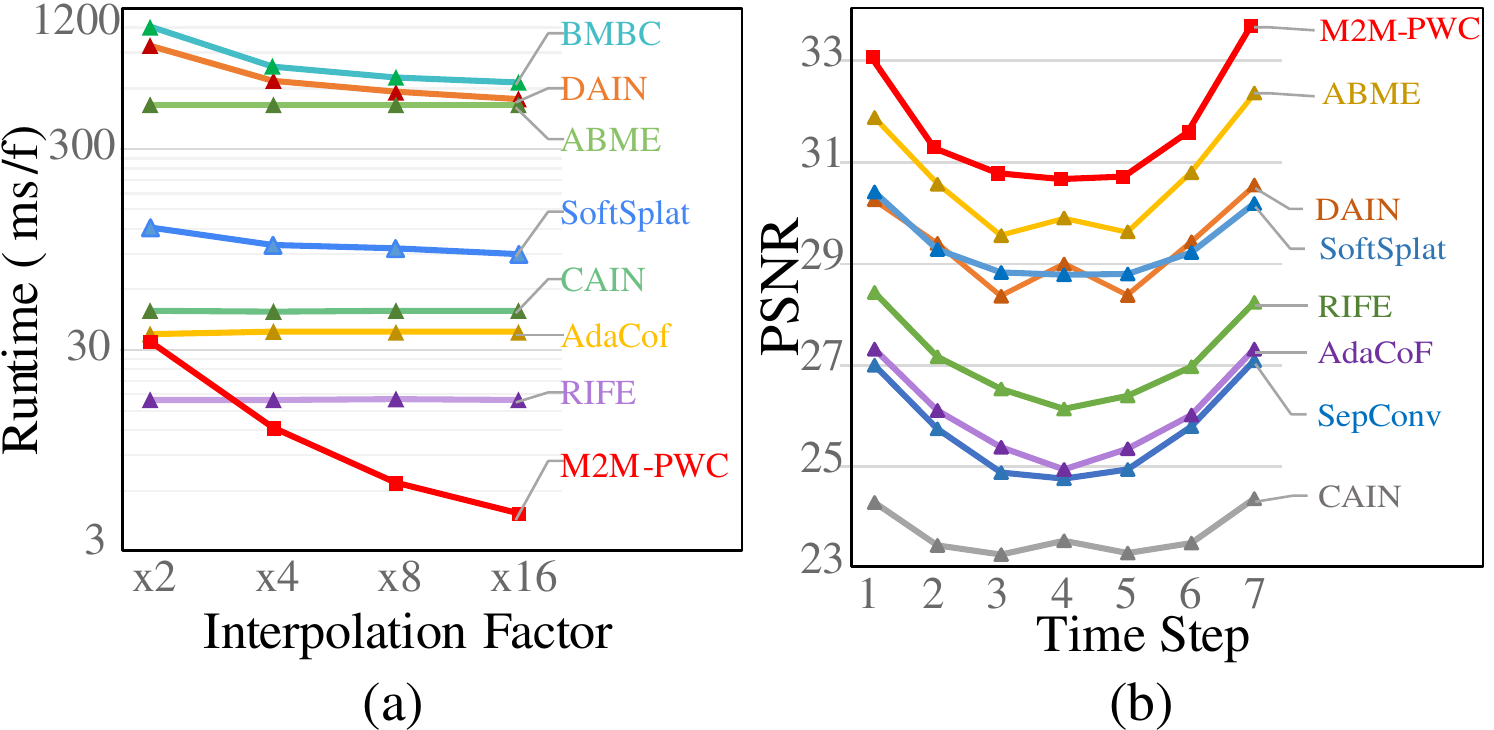}
    \vspace{-0.6cm}
    \caption{\small{Evaluating multi-frame interpolation. (a) Runtime in logarithmic scale for interpolating 640$\times$480 video frames with different interpolation factors. (b) Per-frame accuracy for $\times$8 interpolation on X-TEXT(2K). Best viewed in color.}}
    \vspace{-0.3cm}
    \label{fig:fr_vs_time}
\end{figure}

%% file: tables/benchmark_xvfi.tex
\begin{table}[t]
\small
\begin{tabular}{lp{0.73cm}p{0.73cm}p{0.73cm}p{0.73cm}c}
\toprule[1pt] 
                                            & \multicolumn{2}{c}{X-TEST(4K)}        & \multicolumn{2}{c}{X-TEST(2K)} &Runtime\\
                                                \cmidrule(lr){2-3}\cmidrule(lr){4-6}
                                            & PSNR              & SSIM              & PSNR  & SSIM &(ms/f)\\
\midrule
SepConv~\cite{niklaus2017sepconv}           &23.94              &.794               &25.70  &.800      &~693         \\
DAIN~\cite{bao2019depth}                    &26.78$^{*}$        &.807$^{*}$         &29.33  &.910      &3132         \\
CAIN~\cite{choi2020channel}                 &22.51              &.775               &23.62  &.773      &~287         \\
AdaCoF~\cite{lee2020adacof}                 &23.90              &.727               &26.03  &.778      &~234         \\
SoftSplat~\cite{niklaus2020softmax}         &25.48              &.725               &29.73  &.824      &~318        \\
RIFE~\cite{huang2020rife}                   &24.67              &.797               &27.49  &.806      &~104         \\
ABME~\cite{park2021asymmetric}              &30.16$^{*}$        &.879$^{*}$         &30.65  &.912      &~2904         \\
XVFI$^{\dag}$~\cite{sim2021xvfi}            &30.12              &.870               &30.85  &\underline{.913}      &~203         \\
\midrule
\textbf{M2M-PWC}                 &\textbf{30.81}        &\textbf{.912}   &\textbf{32.07}\textbf{}  &\textbf{0.923}          &~~\underline{44}\\
\textbf{M2M-DIS}                 &\underline{30.18}     &\underline{.909}      &\underline{30.98}  &0.912          &~~\textbf{39}\\
\bottomrule
\end{tabular}
\vspace{-0.2cm}
\caption{\small{Quantitative results for $\times$8 interpolation on the X-TEST dataset. $^\dag$ indicates model trained with X-TRAIN. $^*$ indicates the numbers are copied from~\cite{park2021asymmetric}. All the run-times are measured on X-TEST(2K).}}
\vspace{-0.6cm}
\label{tab:benchmark_xvfi}
\end{table}

%% file: tables/ablation.tex
\begin{table}[]
\centering
\begin{tabular}{p{0.77cm}p{0.77cm}p{0.77cm}p{0.77cm}|cc}
\toprule[1pt] 
                          {MRN}  &{JFE} &{LFM}       &{RS}   &{PWC-Net} &{DISFlow} \\ 
    \hline
               &               &                   &                               &33.97    &31.93\\ 
    \hline
  ~~~$\checkmark$ &               &                   &                               &34.94    &34.32\\ 
  ~~~$\checkmark$ &~~~$\checkmark$   &                   &                               &35.09    &34.59\\ 
  ~~~$\checkmark$ &               &~~~$\checkmark$       &                               &35.07    &34.51\\ 
  \hline
  ~~~$\checkmark$ &~~~$\checkmark$   &~~~$\checkmark$       &                               &35.15    &34.78\\ 
  ~~~$\checkmark$ &~~~$\checkmark$   &~~~$\checkmark$       &~~~$\checkmark$                   &35.24    &34.93\\ 
\bottomrule
\end{tabular}
\vspace{-0.3cm}
\caption{Ablative experiments (in PSNR) on Vimeo90K with different initial flow methods. ``MRN'' denotes the motion refinement network, ``JFE'' refers to the joint flow encoding module in MRN, ``LFM'' is the low-rank feature modulation, and ``RS'' denotes the reliability score in the fusion step that synthesizes the output.}
\vspace{-0.2cm}
\label{tab:ablation}
\end{table}

%% file: tables/branch.tex
\begin{table}[]
\centering
\begin{tabular}{cc|cccc}
\toprule[1pt] 
                                &                       &$N$=1      &$N$=2      &$N$=4  &$N$=8\\ 
    \hline
    \multirow{2}{*}{\small{PWC-Net} }   &\small{PSNR}           &35.24     &35.35       &35.40  &35.39\\ 
                                &\small{Runtime}        &16        &16          &17     &20       \\ 
                                 \hline
    \multirow{2}{*}{\small{DISFlow} }   &\small{PSNR}           &34.93     &34.98       &35.06  &35.07\\ 
                                &\small{Runtime}        &12        &12          &13     &15\\ 
\bottomrule[1pt]
\end{tabular}
\vspace{-0.3cm}
\caption{Analyzing the impact of the number of sub-motion vectors for each pixel in our many-to-many splatting on Vimeo90K, with two different initial flow estimators.}
\vspace{-0.4cm}
\label{tab:branch}
\end{table}

\iffalse
\begin{table}[]
\centering
\small
\begin{tabular}{cp{0.65cm}p{0.65cm}p{1cm}p{0.65cm}p{0.65cm}p{1cm}}
    \toprule[1pt] 
            &\multicolumn{3}{c}{PWC-Net}   &\multicolumn{3}{c}{DISFlow}\\
    \cmidrule(lr){2-4} \cmidrule(lr){5-7}
     \textit{N=}       &\small{PSNR} &\small{SSIM} &\small{Runtime} &\small{PSNR}  &\small{SSIM} &\small{Runtime}\\
    \midrule
    1     &35.24  &&16   &34.93  &&12\\           
    2     &35.35  &&16   &34.98  &&12\\           
    4     &35.41  &&17   &35.06  &&13\\         
    8     &35.39  &&20   &35.07  &&15\\           
    \bottomrule[1pt]
\end{tabular}
\vspace{-0.1cm}
\caption{Analyzing the impact of the number of sub-motion vectors for each pixel in our many-to-many splatting on Vimeo90K, with two different initial flow estimators.}
\vspace{-0.2cm}
\label{tab:branch}
\end{table}

\fi

%% file: figures/xtest_vis.tex
\begin{figure*}[t]
    \centering
    \includegraphics[width=1\linewidth]{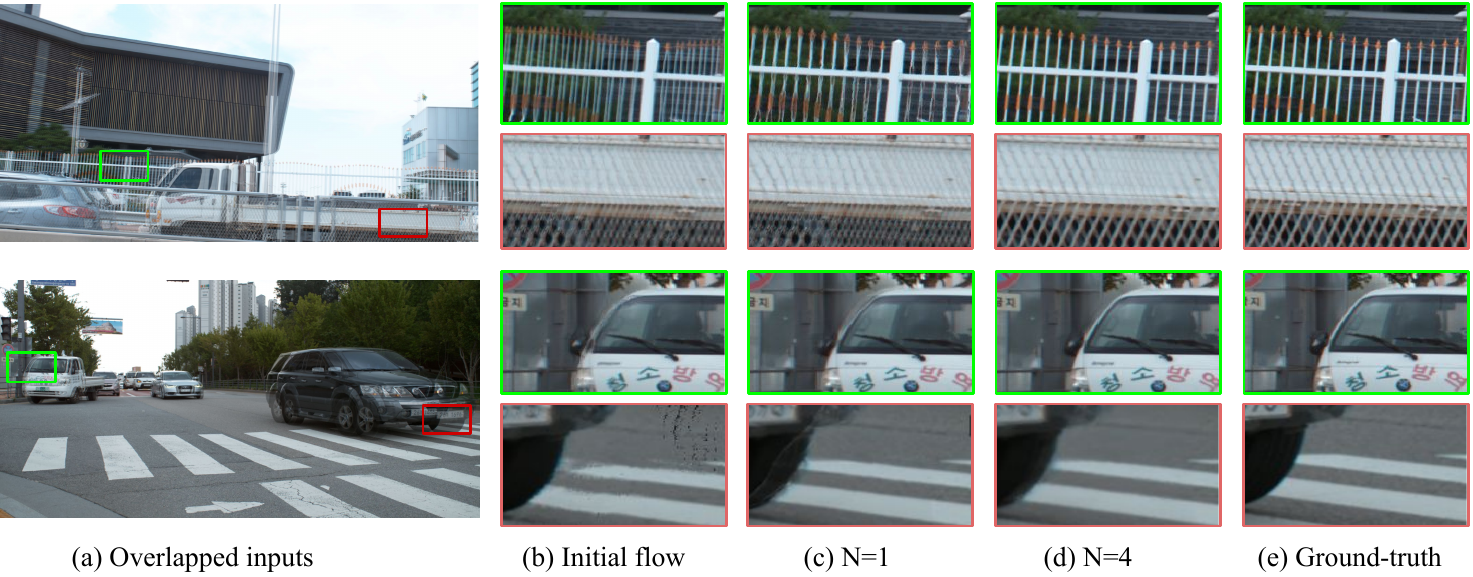}
    \vspace{-0.7cm}
    \caption{\small{Comparison between many-to-one splatting and  many-to-many splatting. Given the input frames (a), M2O splatting with initial flow (b) or single refined sub-motion vector (c) results in undesired visual artifacts for regions with complex motion. In comparison, our proposed M2M splatting with four sub-motion vectors (b) can interpolate with much higher quality.}}
    \vspace{-0.4cm}
    \label{fig:xtest_vis}
\end{figure*}

%% file: tables/resolution.tex
\begin{table}[t]
\centering
\small
\begin{tabular}{p{0.2cm}cccccc}
    \toprule[1pt] 
    &\textit{R=}     &\footnotesize{Xiph-2K}  &\footnotesize{Xiph-``4k''}  &\footnotesize{X-TEST(2K)}  &\footnotesize{X-TEST(4K)} \\
    \midrule
    \multirow{4}{*}{\rotatebox{90}{PWC-Net} }&1   &36.15              &32.94           &28.35             &24.85\\           
                                             &2   &\textbf{36.45}     &33.76           &31.00             &27.08\\           
                                             &4   &36.36              &\textbf{33.93}  &\textbf{32.07}    &29.65\\           
                                             &8   &35.74              &33.75           &31.65      &\textbf{30.81}\\
    \midrule
    \multirow{4}{*}{\rotatebox{90}{DISFlow} }&1   &\textbf{36.14}     &\textbf{33.25}  &31.03             &\textbf{30.18}\\           
                                             &2   &36.05              &33.18           &\textbf{31.18}    &30.06\\           
                                             &4   &35.73              &32.94           &30.54             &29.68\\           
                                             &8   &35.13              &32.29           &29.49             &28.66\\ 
    \bottomrule[1pt]
\end{tabular}
\vspace{-0.2cm}
\caption{Impact of the resolution at which the initial optical flow estimator is applied on. ``R'' is the down-sampling factor.}
\vspace{-0.6cm}
\label{tab:resolution}
\end{table}

%% file: figures/ave_hole.tex
\begin{figure}\vspace{-0.2cm}
    \centering
    \includegraphics[width=0.92\linewidth]{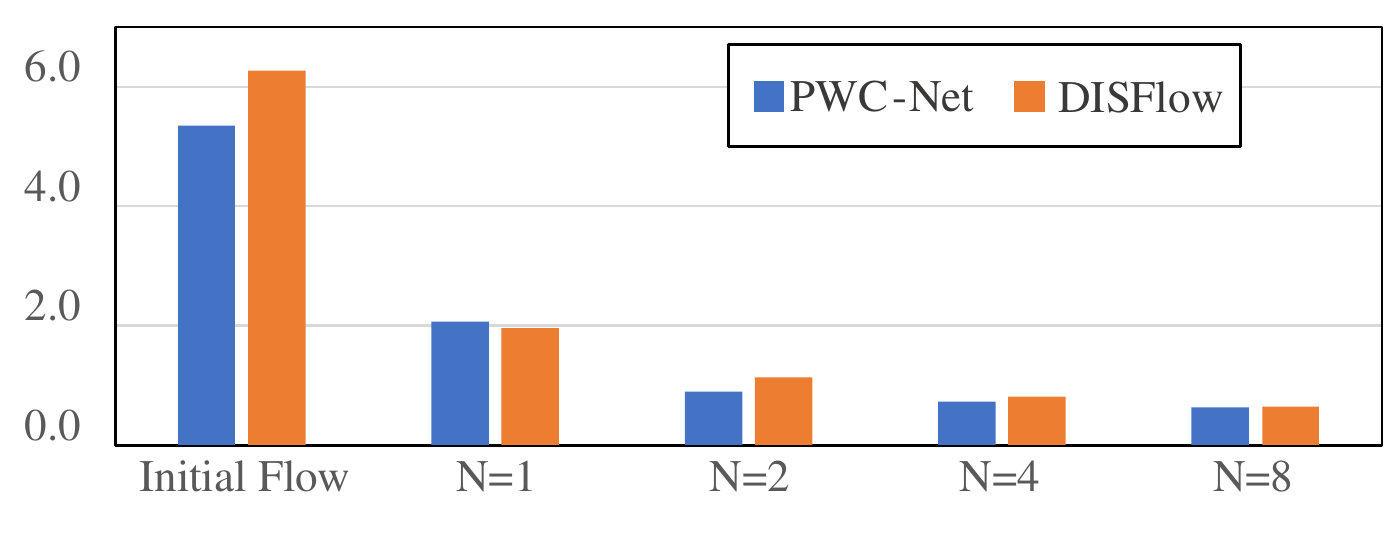}
    \vspace{-0.5cm}
    \caption{\small{Analysis of number of remaining holes (in pixels) versus the number of sub-motion vectors in many-to-many splatting.}}
    \vspace{-0.5cm}
    \label{fig:ave_hole}
\end{figure}